\pdfoutput=1

\documentclass[11pt]{article}

\usepackage{EMNLP2023}

\usepackage{times}
\usepackage{latexsym}

\usepackage[T1]{fontenc}

\usepackage[utf8]{inputenc}

\usepackage{microtype}
\usepackage{booktabs}
\usepackage{float}
\usepackage{amsmath}
\usepackage{graphicx}
\usepackage{multirow}
\usepackage{caption}
\usepackage{xcolor}
\usepackage{epsfig}
\usepackage{graphicx}
\usepackage{wrapfig}
\usepackage{subfig}
\usepackage[belowskip=0pt,aboveskip=0pt,font=small]{caption}
\usepackage{todonotes}
\usepackage{pifont}
\usepackage{dingbat}
\presetkeys{todonotes}{inline}{}

\usepackage{inconsolata}
\usepackage{tablefootnote}
\newcommand{\apx}{{\approx}}

\title{An Empirical Evaluation of Encoder Architectures \\for Fast  Real-Time Long Conversational Understanding}

\author{Annamalai Senthilnathan, Kristjan Arumae, Mohammed Khalilia, \\
{\bf Zhengzheng Xing \& Aaron R. Colak }\\
  Qualtrics Inc.\\
  Seattle, USA \\
  \texttt{\{asenthilnathan, kristjana, mkhalilia, zxing, aaronrc\}@qualtrics.com}}
  
\begin{document}
\maketitle
\begin{abstract}
Analyzing long text data such as customer call transcripts is a cost-intensive and tedious task. 
Machine learning methods, namely Transformers, are leveraged to model agent-customer interactions.
Unfortunately, Transformers adhere to fixed-length architectures and their self-attention mechanism scales quadratically with input length.
Such limitations make it challenging to leverage traditional Transformers for long sequence tasks, such as conversational understanding, especially in real-time use cases. 
In this paper we explore and evaluate recently proposed efficient Transformer variants (e.g. Performer, Reformer) and a CNN-based architecture for real-time and near real-time long conversational understanding tasks.
We show that CNN-based models are dynamic, $\apx$2.6$\times$ faster to train, $\apx$80$\%$ faster inference and $\apx$72$\%$ more memory efficient compared to Transformers on average. 
Additionally, we evaluate the CNN model using the Long Range Arena benchmark to demonstrate competitiveness in general long document analysis.
\end{abstract}

\section{Introduction}

Many organizations depend on customer support calls to guide, support, and deliver a better customer experience.
Analysis of agent-customer calls provides actionable insights to improve business processes around resolving issues with empathy and speed \citep{campos2022compressing}.
This analysis entails addressing a unique set of natural language understanding problems to understand calls in \textit{real time} and \textit{at scale}. 
Our \textit{use case} is to model agent-customer behaviors and call outcomes at various granularities, i.e. conversation and utterance level.
Since such behaviors may not be localized and can span across an entire conversation, we must leverage methodologies than can model the entire conversation, which are generally longer in nature (in the order of thousands of words).

To do this we require an encoder to dynamically scale to arbitrary input length while allowing the latent semantic representation to capture both global and local understanding to accurately predict at different granularities. 
Furthermore, high call volumes (reaching tens of millions monthly) make it prohibitive to use computationally expensive models at scale and many systems need to operate in real time.
Therefore, for real-time inference, models need to be computationally faster and have a smaller memory footprint.

In the past several years, the NLP community has seen Transformers \citep{vaswani2017attention} become the de facto encoder for a multitude of natural language understanding tasks \citep{devlin-etal-2019-bert, brown2020language, raffel2020exploring, liu2019roberta}. 
Pretrained Transformers for NLP tasks employ a self-attention mechanism to capture both short and long-range correlations between tokens. 
They generally succumb to fixed-length architectures (commonly a $512$ sequence length\footnote{We acknowledge that recent LLM developments \citep{openai2023gpt4} frequently go beyond the 512 limit. These models are, however, inexplicably large and not fit for the real-time uses cases we discuss and are therefore beyond the scope of this study.}) and rely on a self-attention mechanism with a quadratic time and memory complexity. 
Thus, scaling traditional Transformers for long conversational inputs can be costly, and may suffer loss of information when inputs are truncated to fit into fixed length architectures.

Various efficient Transformer architectures have been proposed that are high performing, computationally efficient, and can handle longer, albeit fixed length, sequences \citep{tay2020efficient}.
CNN-based architectures are similarly light, fast and with the added benefit of being scalable to inputs of any length. 
There is evidence that CNN models are not only efficient, but also at par in performance with Transformers in various natural language tasks and applications \cite{wu2018pay, tay-etal-2021-pretrained}.

Our goal is to search for an efficient architecture that reduces overall training costs, latency, and provides the flexibility to deploy on less costly hardware, all the while preserving model quality for our use case. 
For this, we consider and evaluate both efficient Transformer and CNN-based architectures.
Our contributions are as follows:
\begin{itemize}
    \item
    We provide evidence that a CNN-based architecture performs competitively against efficient Transformer counterparts in the  conversational domain, while achieving impressive cost benefits, reducing training time, memory, and latency.
    \item
    For a more generalized evaluation, we further evaluate the efficacy of the CNN-based model and analyze its trade-offs against efficient Transformer variants \citep{tay2020efficient} on long sequence tasks from the publicly available Long Range Arena (LRA) benchmark \citep{tay2021long}.
\end{itemize}

\begin{figure*}[t]
    \centering
    \includegraphics[scale=0.2559]{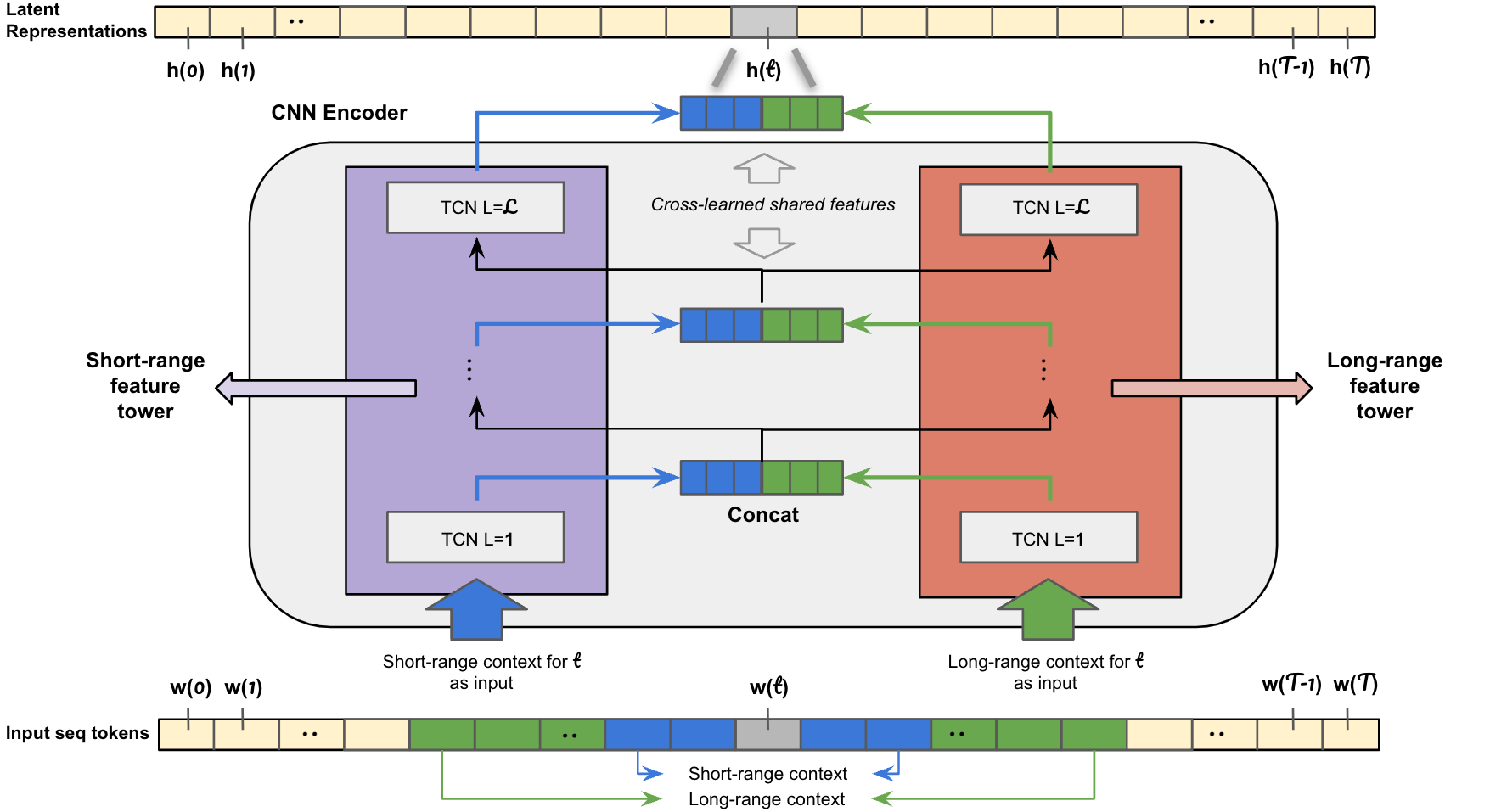}
    \centering
    \caption{Our CNN model with TCN at its core. Here, TCN learns from bi-directional context, \textit{\textbf{t}} is the current token and \textit{\textbf{T}} is the max sequence length of the conversation, \textit{\textbf{w}} is the word embedding \textit{(n-dim vector each)}, \textit{\textbf{h}} is the latent representation \textit{(m-dim vector each)} of the respective token, and \textit{\textbf{L}} denotes the number of layers.}
    \label{fig:model}
\end{figure*}

\section{Related Work}

\subsection{Efficient Transformers} \label{eff_trans_rel_work}

Multiple Transformer variants have been proposed to improve efficiency and handle longer sequences. 
A sparse factorization of the attention matrix was used for reducing the overall complexity from quadratic to $\mathcal{O}(n\sqrt{n})$ for generative modeling of long sequences \citep{child2019generating}. 
\citet{Kitaev2020Reformer} introduced the Reformer model, further reducing the complexity to $\mathcal{O}(n\log{}n)$ via locality-sensitive-hashing. 
A number of linear complexity Transformers have been proposed including Linformer \citep{wang2020linformer}, Longformer \citep{beltagy2020longformer}, BigBird \citep{zaheer2020big}, Nystr{\"o}mformer \citep{xiong2021nystromformer}, and Performer \citep{choromanski2021rethinking} to name a few. 
We refer the reader to a survey on efficient Transformers \citep{tay2020efficient} that succinctly captures the techniques these models use.


\subsection{CNN-based models}

CNN-based models have long been proposed for sequence modeling and NLP applications \citep{kim-2014-convolutional, bai2018empirical, kalchbrenner2016neural}. 
They are lightweight, fast and accurate, especially for text classification. 
CNN models with gated linear units \citep{dauphin2017language} and residual connections \citep{he2016deep} achieved impressive results over Long-Short Term Memory (LSTM) \citep{hochreiter1997long} in seq2seq modeling \citep{gehring2017convolutional}. 
A simple causal CNN architecture demonstrated holding longer memory effectively, outperformed recurrent networks in a diverse range of tasks \citep{bai2018empirical}.
LightConv and DynamicConv \citep{wu2018pay} are CNN-based models for seq2seq learning which outperform Transformers on a series of canonical benchmarks such as machine translation and language modeling.  
\citet{tay-etal-2021-pretrained} observed that pretrained CNN models are competitive with pretrained Transformers in a set of NLP tasks such as sentiment classification, toxicity detection, news classification, and query understanding. 
However, they also noted that purely CNN-based models may not perform well where a cross-attention inductive bias is essential.

Motivated by the need for more efficient architectures, we evaluate both CNN-based models and efficient Transformers and analyze their trade-offs in conversational settings. 

\section{Methodology} \label{sec:our_model}
Our CNN-based architecture, shown in Figure~\ref{fig:model}, is composed of two stacks or towers of Temporal Convolutional Networks (TCN), originally proposed by \citet{bai2018empirical}. 
We employ these two towers together to allow us to accurately predict at different granularities. 
We are inspired by TCN's simplicity and combinations of the best practices in contemporary convolutional modeling, namely dilated convolutions \citep{yu2015multi, oord2016wavenet} and residual connections \citep{7780459}.

We tackle two main challenges in long conversation modeling with this arrangement. 
First, we want to leverage past and future information, and second, we want to capture near and distant contextual signals. 
Unlike the original TCN architecture that only learns from past information, we tune the padding factor to leverage the bi-directional context (past and future) within a conversation or document, thereby addressing the first challenge.

To capture the long and short range context, we tune the reception field range such that one tower attends to the short range, and the other attends to the long-range contextual features. 
We concatenate the outputs of each layer and feed that as input to the subsequent layer of each tower to induce cross-learning between features. 
We use global max-pooling for conversation level and local max-pooling (pooling utterance-specific latent representations only) for utterance level predictions.

We follow the same dilation factor and residual connection strategies as described in \citet{bai2018empirical}. 
In this paper we refer to this architecture as the \textit{CNN model}, unless otherwise specified.




\section{Tasks and Experiments}
In this section we describe the tasks, data, and experimental setup of both Conversational Understanding and the LRA benchmark.

\subsection{Conversational Understanding}
\subsubsection{Tasks}
We consider two tasks that require understanding at different granularities:
(i) conversation level identification of the overall agent-customer behaviors, and
(ii) utterance level evidence detection for the same behaviors.
For experimentation and evaluation both public and proprietary datasets will follow similar task design.

\subsubsection{Datasets} \label{sec:datasets}
\noindent\textbf{Repurposed Action-Based Conversation Data (Repur. ABCD):}
The original Action-Based Conversation Dataset (ABCD) is a publicly available task-oriented conversational dataset \cite{chen-etal-2021-action}. 
It is a synthetic human-to-human conversational dataset with over 10K dialogues containing 55 distinct \textit{customer intents} requiring unique \textit{sequences of actions} constrained by policy-driven agent guidelines to achieve task success. 

At a high level, the agent guidelines can be decomposed into three levels of annotation: \textit{Flows, Subflows} and \textit{Actions}. 
Flows describe the overall issue discussed in the call.
Actions describe the most granular activities that the agent can perform at a turn level. 
Subflows, an intermediate level between high-level flows and granular actions, define a set of \textit{customer intents} that are to be addressed by \textit{sequences of actions} during the course of the call.
For the full ontology of the agent guidelines, we refer the reader to \citet{chen-etal-2021-action}.

\citet{chen-etal-2021-action} proposed two tasks, \textit{Action State Tracking} (AST) and \textit{Cascading Dialogue Success} (CDS).
AST is a time-series action prediction task at turn level, where the model predicts the next action(s) the agent should take based on the conversation history, predicated on agent guidelines.
CDS extends this notion across the entire conversation. 
It measures the model's ability to understand actions over consecutive turns, recommends the next utterance or to end the conversation.

We re-purpose the ABCD dataset to be similar in granularity to our desired tasks\footnote{Note that these repurposed tasks are different from AST and CDS, thereby making our results non-comparable with the results of \citet{chen-etal-2021-action}. 
More details on our data conversion can be found in Appendix \ref{subsec:data_trans}.}. 
The \textit{Flow} labels in our use case are used at the conversation level (\textit{call issue identification}) and modeled as a \textit{multi-class classification task} since there is only one issue discussed in a given call.

The \textit{Action} labels correspond to utterance level labels (\textit{agent action prediction}) by tagging the most recent past utterance with the action labels of the given turn. 
This task is modeled as \textit{multi-label classification}, since any number of actions may be performed at a given point in the conversation. 


\begin{table}[t]
    \centering
    \resizebox{\columnwidth}{!}{
    \begin{tabular}{lrr}
    \toprule
    Dataset Stats & Repur. ABCD & Proprietary \\
    \midrule
    \#Conversations & 10,042 & 1,030 \\
    \#Utterances & 174,691 & 83,205 \\
    \#Conv. level labels & 10 & 5 \\
    \#Utt. level labels & 30 & 20 \\
    Avg. seq. length ($\pm${sd}) & 252 ($\pm${92}) & 1,112 ($\pm${878}) \\
    Max. seq. length & 996 & 4,056 \\
    \bottomrule
    \end{tabular}}
    \caption{Statistics of the Repurposed ABCD and our proprietary dataset. Sequence length refers to the total number of Byte-level Byte-Pair encoding (BBPE) tokens (using the RoBERTa Tokenizer). \textit{sd} is standard deviation}
    \label{table:data_stats}
\end{table}

\noindent\textbf{Proprietary Conversation Data:}
This corpus consists of conversations collected from real agent-customer interactions across a range of industries such as insurance, retail, finance, and healthcare.
We aim to understand these calls at different granularities.
For instance, the agent's overall knowledge or understanding of the customer's issue during the call is pivotal in an effective resolution.
We can also detect behavior at the utterance level, which can serve as evidence to the agent's overall performance in the call. 
%
In this dataset both our conversation and utterance level tasks are modeled as multi-label classification.

Table \ref{table:data_stats} provides statistics of both the Repurposed ABCD dataset and our proprietary dataset. 
For corpus label distributions please refer to Appendix \ref{subsec:appx_datasets}.

\subsubsection{Experimental setup}
We perform experiments to identify an efficient architecture that gives best results in terms of cost and performance. 
We capture both performance and efficiency metrics for all the models.
To leverage semantic understanding across the granularity of the tasks and to more realistically\footnote{This can be considered realistic due to the efficiency of maintaining a single model over many, as well as the reduced latency of a single model call.} model the conversation and utterance level labels we approach this jointly from a multi-task learning (MTL) perspective.
Our effort focuses on MTL, however we also train single task learning (STL) models to provide additional perspective on model performance. 
STL related performance can be found in Appendix \ref{subsec:appx_res_std}.

\noindent\textbf{Models:}
We consider the following efficient Transformer architecture variants: Longformer, the only pretrained model \citep{beltagy2020longformer}, Reformer \citep{Kitaev2020Reformer}, Performer \citep{choromanski2021rethinking}, and Nystr{\"o}mformer \citep{xiong2021nystromformer}. 

\begin{table}[h]
    \centering
    \resizebox{\columnwidth}{!}{
    \begin{tabular}{lrrr}
    \toprule
    Model & \#Params & \#Layers & Pretrained \\
    \midrule
    Longformer & 149 M & 12 & \checkmark \\
    Nystr{\"o}mformer & 107 M & 6 & \ding{55} \\
    Performer & 115 M & 6 & \ding{55} \\
    Reformer & 105 M & 12 & \ding{55} \\
    \midrule
    \textit{CNN}\textsubscript{Large} \textit{(ours)} & 113 M & 4 & \ding{55} \\
    \textit{CNN}\textsubscript{Small} \textit{(ours)} & 46 M & 4 & \ding{55} \\
    \bottomrule
    \end{tabular}}
    \caption{Overview of the models}
    \label{table:model_params}
\end{table}

\noindent\textbf{Implementation details:}
Although Reformer, Performer, Nystr{\"o}mformer can be parameterized to ingest longer sequences, we have set their maximum sequence length to $4,096$, to match the Longformer’s length. 
We maintain a comparable number of parameters across all models ($\pm{10\%}$ relative difference), with exception to the Longformer, as shown in Table \ref{table:model_params}.
While the Longformer is pretrained and includes its learned word embeddings, the other models, including CNN, use pretrained RoBERTa word embeddings \citep{liu2019roberta,radford2019language}. 
To demonstrate the parameter efficiency of the CNN architecture we evaluate a lighter version of our CNN model, \textit{CNN}\textsubscript{Small}. 
For additional details like differences between \textit{CNN}\textsubscript{Large} vs. \textit{CNN}\textsubscript{Small} see Appendix \ref{subsec:appx_impl_details}.

\noindent\textbf{Input Format:}
The input to all models are full length conversations. 
At an utterance level, we add speaker identifiers as an enrichment in the form of special tokens. 
We only use conversations\footnote{Across both datasets, we had $40$ conversations that were $> 4,096$ token length. 
Table \ref{table:data_stats} excludes these.} whose max sequence length is $\leq 4,096$ tokens and split the data into train ($80\%$), validation ($10\%$) and test ($10\%$) sets. 

\noindent\textbf{Metrics and Hardware:}
We use the AWS SageMaker accelerated computing instance\footnote{\url{https://aws.amazon.com/sagemaker/pricing/}} \textit{ml.g5.12xlarge} for our experiments.
For inference latency, we host models on \textit{ml.g4dn.2xlarge}, 
and calculate the inference time in seconds for batches of size $8$\footnote{Longformer and Nystr{\"o}mformer resulted in out-of-memory error at inference for a batch size of $8$ conversations. For these, we showcase the aggregated latency from smaller batch sizes for comparison.}.
We measure model performance by macro-averaged F1 on the test set, training speed by steps per second, and the memory footprint by the maximum GPU memory usage (GBs) as observed by AWS CloudWatch\footnote{\url{https://docs.aws.amazon.com/sagemaker/latest/dg/monitoring-cloudwatch.html}}. 

\noindent\textbf{Optimization and Hyperparameters:}
Given the large space of hyperparameters and their expensive tuning cost, we follow the architectural choices for the efficient Transformers from their respective publications. 
For other hyperparameters we follow \citet{xiong2021nystromformer}, which uses the Adam optimizer \citep{kingma2014adam}, learning rate $\eta=1.0\mathrm{e}{-4}$, $\beta\textsubscript{1}=0.9$, $\beta\textsubscript{2}=0.999$, $L_2$ weight decay of $0.01$, learning rate warm-up to $30\%$ of the steps in a given experiment and linear learning rate decay to update model parameters. 
For CNN models we use the SGD optimizer \citep{ruder2016overview, loshchilov2017sgdr}, which performed better than Adam, a one-cycle learning rate \citep{smith2019super} scheduler and a maximum $\eta$ of $0.01$. 
We observed that baseline models performed worse when using SGD over Adam. 
For more details on the hyperparameters used, please refer to Appendix~\ref{subsec:appx_impl_details}.

\subsection{Long Range Arena (LRA) Benchmark}
To provide a well-rounded performance comparison we benchmark our CNN model in an established long-input sequence modeling testing suite, the LRA benchmark \citep{tay2021long}. 
All experiments were conducted using AWS \textit{ml.g5.12xlarge} instances.

\subsubsection{Tasks and Data}
The LRA benchmark has several tasks, out of which we focus on the three text-based evaluations: (i) byte-level text classification of IMDb reviews \citep{maas2011learning},
(ii) ListOps \citep{nangia-bowman-2018-listops}, and
(iii) byte-level document retrieval of the ACL Anthology Network \citep{radev2013acl}.

\begin{table}[t]
    \centering
    \resizebox{\columnwidth}{!}{
    \begin{tabular}{ll|*2r|*3c}
    \toprule
    Data & Model & Conv. & Utt. & Train & Mem & Latency \\
    \midrule
    \quad \multirow{5}{*}{\rotatebox{90}{Repur. ABCD}}
    & Longformer & 91.8 & 22.2 & 0.3 & 58.1 & 0.8 \\
    & Nystr{\"o}mformer & 88.6 & 33.6 & 0.3 & 55.4 & 0.7 \\
    & Performer & 76.5 & 21.7 & 0.8 & 39.6 & 0.3 \\
    & Reformer & 41.4 & 4.5 & 1.5 & 14.5 & 0.2 \\
    \cmidrule{2-7}
    & CNN\textsubscript{Large} \textit{(ours)} & \textbf{94.9} & \textbf{55.6} & 0.8 & 17.9 & 0.3 \\
    & CNN\textsubscript{Small} \textit{(ours)} & 94.0 & 52.8 & \textbf{1.9} & \textbf{11.6} & \textbf{0.1} \\
    \midrule
    \quad \multirow{5}{*}{\rotatebox{90}{Proprietary}}
    & Longformer & 55.6 & 9.4 & 0.3 & 58.1 & 1.9 \\
    & Nystr{\"o}mformer & 46.2 & 33.7 & 0.2 & 55.4 & 1.7 \\
    & Performer & 48.1 & 38.3 & 0.6 & 39.8 & 0.3 \\
    & Reformer & 43.3 & 18.5 & \textbf{1.2} & 17.6 & 0.2 \\
    \cmidrule{2-7}
    & CNN\textsubscript{Large} \textit{(ours)} & \textbf{57.9} & 49.8 & 0.5 & 18.8 & 0.3 \\
    & CNN\textsubscript{Small} \textit{(ours)} & 57.1 & \textbf{52.3} & \textbf{1.2} & \textbf{11.7} & \textbf{0.1}\\
    \bottomrule
    \end{tabular}}
    \caption{Model’s performance (F1) across datasets and tasks (conv. and utt.) trained in MTL paradigm. \textit{Train} is the Training speed. \textit{Mem} is the peak GPU memory used.}
    \label{table:conv_und_results}
\end{table}

\subsubsection{Experimental Setup} \label{subsec:lra_impl_details}
\noindent \textbf{Implementation details:}
The official LRA benchmark codebase\footnote{\url{https://github.com/google-research/long-range-arena}} is implemented in JAX\footnote{\url{https://github.com/google/jax}} and uses Flax\footnote{\url{https://github.com/google/flax}} neural network library.
We, however, follow the \citet{xiong2021nystromformer} PyTorch \citep{NEURIPS2019_bdbca288} implementation.

\noindent \textbf{Baselines:} 
The baseline models we use are the same as in \citet{xiong2021nystromformer} which include the models presented in Table \ref{table:model_params} with the exception of Longformer. The architecture and hyperparameters of the baselines are as discussed in \citet{xiong2021nystromformer} unless otherwise specified.
Please see Appendix \ref{subsec:appx_lra_impl_details} for more details.


\noindent \textbf{Metrics:} We follow the evaluation protocol from \citet{tay2021long} including train/test splits, and report the classification accuracy for each task, as well as the average across all tasks.  
We count the FLOPs (G) of the models using fvcore\footnote{\url{https://github.com/facebookresearch/fvcore}}. 
We compute steps/second to measure training speed and report the relative speed compared to the vanilla self-attention. 
We monitor the maximum GPU memory usage (GBs) as memory footprint, using AWS CloudWatch.

\begin{table*}[t]
    \centering
    \resizebox{\textwidth}{!}{
    \begin{tabular}{l|rccc|rccc|rccc|c}
    \toprule
    & \multicolumn{4}{c|}{Text} & \multicolumn{4}{c|}{ListOps} & \multicolumn{4}{c|}{Retrieval} & Average \\
    & \multicolumn{4}{c|}{(1296 $\pm${893})} & \multicolumn{4}{c|}{(888 $\pm${339})} & \multicolumn{4}{c|}{(3987 $\pm${560})} \\
    \midrule
    Models & Acc. & FLOPs & Tr. speed & Mem. & Acc. & FLOPs & Tr. speed & Mem. & Acc. & FLOPs & Tr. speed & Mem. & Acc. \\
    \midrule
    Full Attention & 65.7 & 4.6 & 1.0$\times$ & 35.8 & \textbf{36.8} & 1.2 & \textbf{1.0}$\times$ & 12.6 & 79.8 & 9.1 & 1.0$\times$ & 70.8 & \textbf{60.8} \\
    \midrule
    Reformer & 65.2 & 0.6 & 1.6$\times$ & 8.5 & 24.9 & 0.3 & 0.8$\times$ & 6.6 & 78.3 & 1.2 & 1.9$\times$ & 11.5 & 56.1 \\
    Performer & 65.1 & 0.8 & 1.8$\times$ & 8.4 & 28.1 & 0.4 & 0.9$\times$ & 6.7 & 79.3 & 1.6 & 2.1$\times$ & 11.4 & 57.5 \\
    Nystr{\"o}mformer & 65.8 & 1.0 & 1.6$\times$ & 7.8 & 31.4 & 0.6 & 0.8$\times$ & 6.4 & \textbf{80.3} & 2.0 & 1.9$\times$ & 7.1 & 59.2 \\
    \midrule
    CNN \textit{(ours)} & \textbf{68.0} & \textbf{0.3} & \textbf{2.1}$\times$ & \textbf{7.2} & 17.8 & \textbf{0.1} & \textbf{1.0}$\times$ & 7.1 & 72.8 & \textbf{1.0} & \textbf{4.4}$\times$ & \textbf{7.4} & 52.9 \\
    \bottomrule
    \end{tabular}}
    \caption{LRA Benchmark results - Model's accuracy, FLOPs count, relative training speed, and max GPU memory usage}
    \label{table:lra_perf}
\end{table*}

\section{Results and Findings}
All our results are averaged across four experiments with different seeds. 

\subsection{Conversational Understanding}
 

\noindent\textbf{Model Performance:}
We show model performance in conversation and utterance level tasks across both datasets.
From Table \ref{table:conv_und_results}, the CNN model is at par or outperforms the efficient Transformers in both tasks.
Given the large label space and highly skewed distribution, all models struggle in the utterance-level task compared to the conversation-level task with the effect being more noticeable in the baseline models.
The standard deviation of the model performance across these experiments can be found in Appendix \ref{subsec:appx_res_std}.

We believe that the self-attention approximation of these efficient Transformer variants generally does a better job at understanding the context at a document or conversation level, but struggles at a more granular level  (i.e., utterance).


\noindent\textbf{Cost Benefits:}
We show the cost of using each model with their training speed, latency and peak GPU memory usage in Table \ref{table:conv_und_results}. 
Compared to efficient Transformers, we observe that 
(i) the CNN model trains faster, and has a lower memory footprint and latency while achieving competitive performance in conversational understanding tasks;
(ii) on average, CNN\textsubscript{Small} is $\apx$2.6$\times$ faster to train and achieves $\apx$72$\%$ reduction in GPU memory usage, $\apx$61$\%$ reduction in parameters and $\apx$80$\%$ reduction in inference latency.


\subsection{LRA Benchmark}
We consolidated our LRA benchmark results in Table \ref{table:lra_perf}. 
Overall, we find the CNN model to be competitive with Transformers, although with some caveats.

In the \textbf{Text} task, which has a high input length variance ($\pm{893}$), the CNN model outperforms all baselines in both performance and cost. 
The CNN model achieves $\apx$93$\%$ reduction in FLOPs, a $\apx$2.1$\times$ increase in training speed, and a $\apx80\%$ reduction in memory compared to the vanilla Transformer. 
In \textbf{ListOps} task, the CNNs inductive bias may not be as effective as Transformers in learning hierarchically structured data, leading to reduced performance.
In \textbf{Retrieval} task, the CNN model performs lower compared to the baselines, however, has significant reduction in training speed, FLOPs, and memory usage. 
We observe that, in this task, the CNN model enjoys greater cost benefits, but with substantial performance degradation.
So, how much cost benefit can the CNN models sacrifice to achieve similar performance to the baselines in the retrieval task? 
To understand this, we perform an ablation study on increasing the reception field range of CNN model to improve document understanding and thereby task performance (Table \ref{table:recep_abl}).

Compared to the vanilla self-attention, we observe that CNN\textsubscript{k=257}, where \textit{k} is the CNN kernel size, performs at par and still achieves $\apx$35$\%$ reduction in FLOPs, 1.4$\times$ increase in speed and an order of magnitude lower memory usage.

\begin{table}[h]
    \centering
    \resizebox{3in}{!}{
    \begin{tabular}{llcrr}
    \toprule
    Model & Acc. & FLOPs & Tr. speed & Mem. \\
    \midrule
    Full Attn. & 79.8 & 9.1 & 1.0$\times$ & 70.8 \\
    \midrule
    CNN\textsubscript{\textit{k=17}} & 72.8 & 1.0 & 4.4$\times$ & 7.4 \\
    CNN\textsubscript{\textit{k=21}} & 74.5 & 1.6 & 4.0$\times$ & 7.4 \\
    CNN\textsubscript{\textit{k=257}} & 79.4 & 5.9 & 1.4$\times$ & 7.6 \\
    \bottomrule
    \end{tabular}}
    \caption{Ablation study. The CNN model for Retrieval task in Table \ref{table:lra_perf} has the setting \textit{k}=17 and we increase it in this study. More details in Appendix \ref{subsec:appx_cnn_hyper}}
    \label{table:recep_abl}
\end{table}


\section{Conclusion}
We have presented empirical evidence that CNN-based models exhibit competitive performance and impressive cost savings in modeling task-oriented conversations while remaining scalable to input length. 
Conservatively, using CNN\textsubscript{Small}, we estimate an average of $\apx58\%$ savings for model development costs while preserving model quality compared to the baselines. 
This savings rate will increase when including the potential to reduce instance capacity and deploy using lower cost hardware.
We estimate that we can train on average $2$-$3$ models in the same time-frame as a single model using one of the efficient Transformers.
Considering that our observation could be highly use case specific, we have also shown that CNN-based models perform competitively against Transformers in a more generic Long Range Arena benchmark. 



\section*{Limitations}
Our preliminary results show the performance and efficiency of both Transformer and CNN-based architectures in conversational understanding tasks. 
We are limited by experimentation and resource usage costs.
Despite the encouraging results, we note that searching for optimal hyperparameters for these models is a non-trivial and costly.
By aggressively tuning hyperparameters for these models, the best performance and the relative order of these models may change.
However, the key takeaway is Transformers may not be the best architecture for all cases and that convolutional models are a competitive alternative in certain NLP tasks and production use cases.
Given the cost constraints we were unable to perform further experimentation.
(i) How our CNN model's performand and efficiency scales when we scale its size to billions of parameters (i.e. competition against the post-ChatGPT wave of LLMs)? 
(ii) Potential performance improvements when pretraining our CNN model and leveraging the pretrained model for downstream tasks.




\section*{Ethics Statement}
We recognize the significance of ethical considerations in conducting our research work, and building conversational systems.
Our efforts to develop efficient models that use less power, fewer resources, will support a sustainable environment \citep{strubell-etal-2019-energy}.
We are mindful of issues such as privacy, bias, discrimination etc., and have mitigated it by removing any and all Personal Identifiable Information (PII) in our conversational data before modeling.
Also, we have taken steps to mitigate such issues in data collection, model design and/or experimental setup. 

\bibliography{main}
\bibliographystyle{acl_natbib}

\appendix
\section{Appendix} \label{sec:appendix}

\subsection{Re-purposing ABCD data} \label{subsec:data_trans}
\begin{figure}[!h]
    \centering
    \includegraphics[scale=0.15]{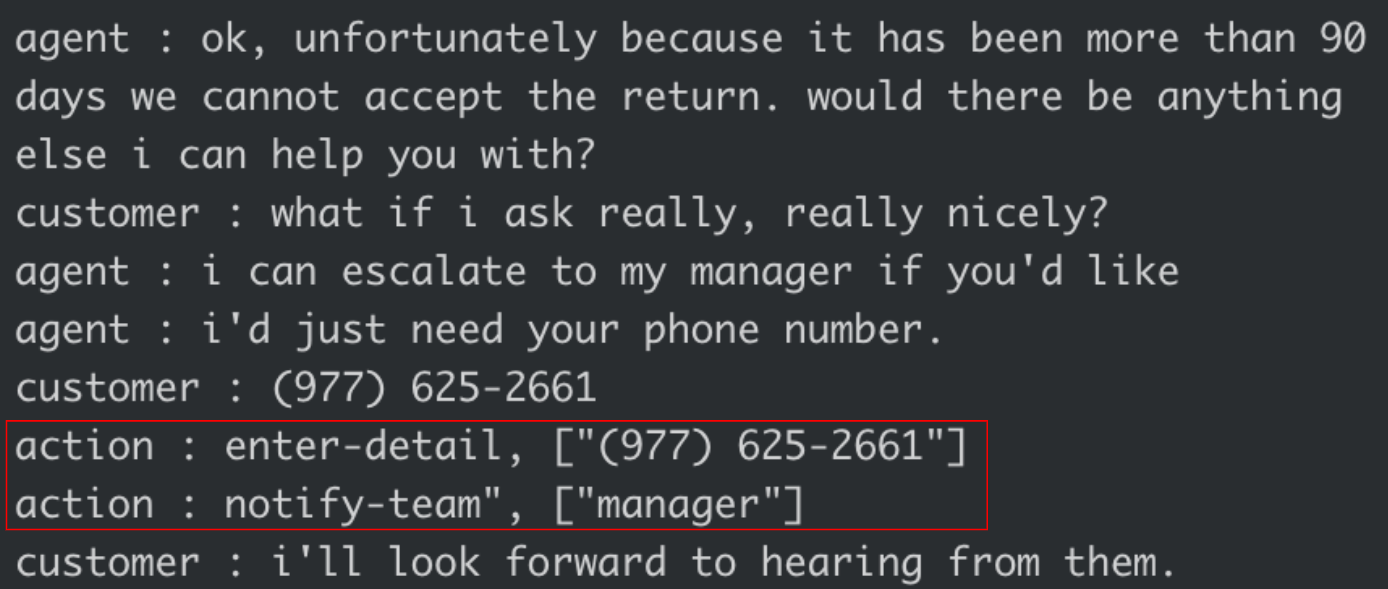}
    \centering
    \caption{An excerpt from the original ABCD dataset, highlighting the action labels and their relevant entities}
    \label{fig:abcd_excerpt}
\end{figure}
Figure \ref{fig:abcd_excerpt} showcases an excerpt of a conversation between agent and customer. 
The \textit{Flow} label for this particular conversation is \textit{"product defect"} in ABCD dataset \citep{chen-etal-2021-action}.
The highlighted box contains the \textit{Action} labels and relevant entities that their chat system is to predict, extract and use the information to prompt the agent at desired times, while abiding to their agent guidelines. 
We want to repurpose data of this sort into tasks similar to our conversational understanding use case.

For our \textbf{utterance level task}, we simplify by focusing only on \textit{Action} labels and not on entities. 
In a conversation, we leverage each of these \textit{Action} labels by adding them as labels to the nearest previous utterance. 
Other utterances will have no actions associated with them.
This kind of \textit{(utterance, actions)} pair makes up the utterance level task data of the Repurposed ABCD dataset and is modeled as multi-label classification task. 
For the example in Figure \ref{fig:abcd_excerpt}, both the action labels \textit{enter-detail} and \textit{notify-team} will be repurposed as utterance labels for the utterance where the customer inputs their phone number. 
For our \textbf{conversation level task}, we simply use the Flow label (customer issue) of each conversation as its conversation level label. 
This \textit{(conversation, flow)} pair makes up the conversation level task data of the Repurposed ABCD dataset and modeled as multi-class classification task.

\begin{table*}[t]
    \centering
    \resizebox{4.6in}{!}{
    \begin{tabular}{ll|*2r|*2r}
    \toprule
    & & \multicolumn{2}{c|}{STL paradigm} & \multicolumn{2}{c}{MTL Paradigm} \\
    \midrule
    Dataset & Model & Conv-task & Utt-task & Conv-task & Utt-task \\
    & & F1 ($\pm${sd}) & F1 ($\pm${sd}) & F1 ($\pm${sd}) & F1 ($\pm${sd}) \\
    \midrule
    \quad \multirow{5}{*}{\rotatebox{90}{Repur. ABCD}}
    & Longformer & 75.0 ($\pm${4.2}) & 22.5 ($\pm${2.0}) & 91.8 ($\pm${0.9}) & 22.2 ($\pm${0.9}) \\
    & Nystr{\"o}mformer & 91.5 ($\pm${0.6}) & 33.7 ($\pm${0.6}) & 88.6 ($\pm${0.5}) & 33.6 ($\pm${0.4}) \\
    & Performer & 86.6 ($\pm${1.3}) & 20.4 ($\pm${0.6}) & 76.5 ($\pm${2.3}) & 21.7 ($\pm${1.1}) \\
    & Reformer & 39.7 ($\pm${7.4}) & 4.3 ($\pm${2.1}) & 41.4 ($\pm${8.7}) & 4.5 ($\pm${2.2}) \\
    \cmidrule{2-6}
    & CNN\textsubscript{Large} \textit{(ours)} & \textbf{94.9} ($\pm${0.3}) & \textbf{55.0} ($\pm${0.6}) & \textbf{94.9} ($\pm${0.3}) & \textbf{55.6} ($\pm${0.8}) \\
    & CNN\textsubscript{Small} \textit{(ours)} & \textbf{94.9} ($\pm${0.4}) & 51.3 ($\pm${0.6}) & 94.0 ($\pm${0.8}) & 52.8 ($\pm${0.5})\\
    
    \midrule
    \quad \multirow{5}{*}{\rotatebox{90}{Proprietary}}
    & Longformer & 49.4 ($\pm${5.2}) & 9.5 ($\pm${0.3}) & 55.6 ($\pm${2.4}) & 9.4 ($\pm${0.3}) \\
    & Nystr{\"o}mformer & 47.9 ($\pm${3.0}) & 26.6 ($\pm${1.0}) & 46.2 ($\pm${3.1}) & 33.7 ($\pm${1.9}) \\
    & Performer & 49.3 ($\pm${3.6}) & 40.1 ($\pm${0.6}) & 48.1 ($\pm${2.3}) & 38.3 ($\pm${0.9}) \\
    & Reformer & 41.6 ($\pm${1.1}) & 17.6 ($\pm${0.8}) & 43.3 ($\pm${4.0}) & 18.5 ($\pm${2.6}) \\
    \cmidrule{2-6}
    & CNN\textsubscript{Large} \textit{(ours)} & \textbf{55.8} ($\pm${1.9}) & \textbf{52.4} ($\pm${2.3}) &  \textbf{57.9} ($\pm${2.6}) & \textbf{49.8} ($\pm${1.9}) \\
    & CNN\textsubscript{Small} \textit{(ours)} & 54.9 ($\pm${3.1}) & 52.0 ($\pm${1.3}) & 57.1 ($\pm${3.1}) & \textbf{52.3} ($\pm${0.5})\\
    \bottomrule
    \end{tabular}}
    \caption{Model’s performance across tasks, datasets and learning paradigm along with its standard deviation.}
    \label{table:appx_conv_und_res_w_std}
\end{table*}

Our tasks are different from AST and CDS tasks proposed in \citet{chen-etal-2021-action}. 
The repurposed data and tasks that we evaluate are much simpler in nature.
Compared to AST, our tasks are not predicated on any account constraints nor agent guidelines. 
Also, we do not extract any relevant entities, nor prompt during the conversational flow. 
We do not use their \textit{Subflow} labels in our tasks and their evaluation of the CDS is different from our conversational task, where our task is a simple multi-class classification task to identify the call issue discussed (\textit{Flow} labels).

\subsection{Conversational Understanding Use Case}
Here we will see additional details around conversational understanding tasks' label distribution, model implementation, hyperparameters and results.
\subsubsection{Label Distribution} \label{subsec:appx_datasets}
\begin{figure*}[h]
\centering
\subfloat[Conversation level label distribution]{\includegraphics[trim=0pt 0pt  0pt 0pt, clip=true, width=0.50\textwidth]{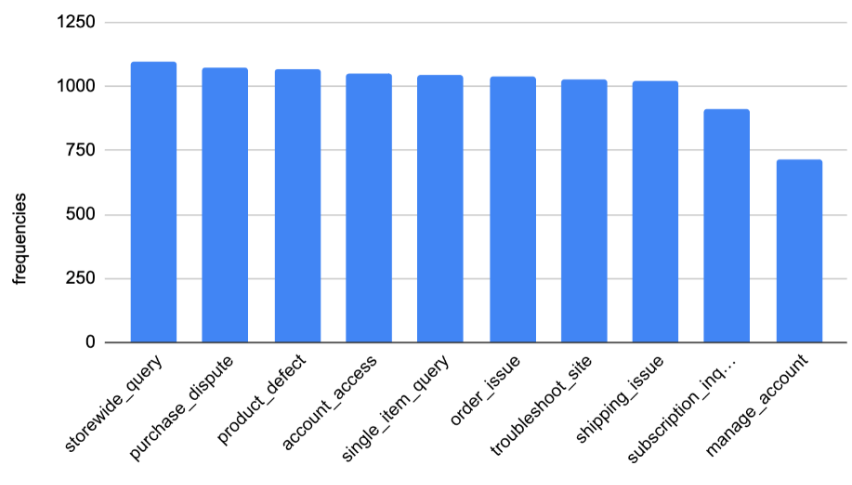}\label{fig:abcd_conv_lbls}}
\subfloat[Utterance level label distribution]{\includegraphics[trim=0pt 0pt  0pt 0pt, clip=true, width=0.50\textwidth]{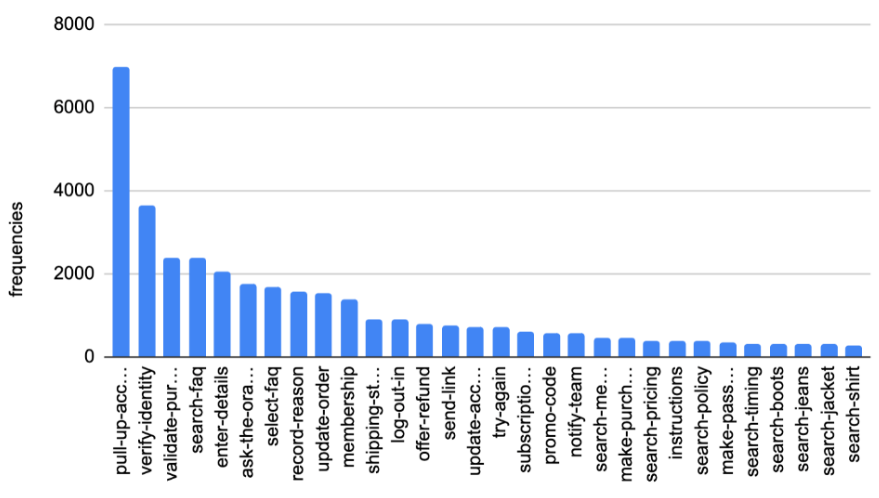}\label{fig:abcd_utt_lbls}}
\caption{Repurposed ABCD dataset's conversation and utterance label distributions}
\label{fig:abcd_data}
\end{figure*}

\begin{figure*}[h]
\centering
\subfloat[Conversation level label distribution]{\includegraphics[trim=0pt 0pt  0pt 0pt, clip=true, width=0.50\textwidth]{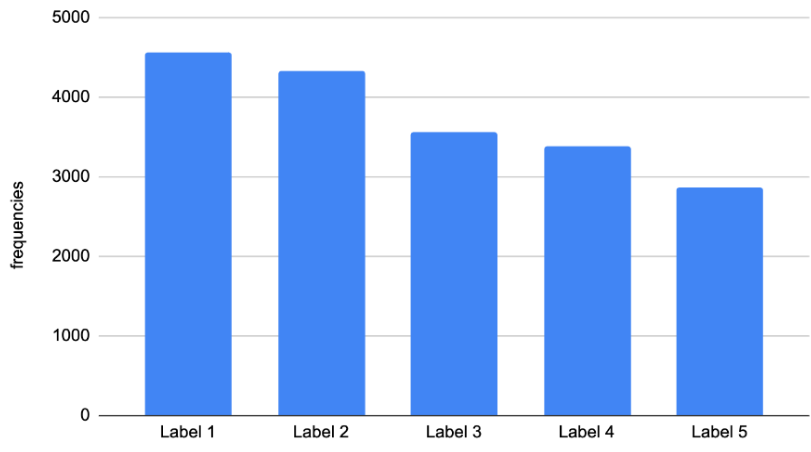}\label{fig:prop_conv_lbls}}
\subfloat[Utterance level label distribution]{\includegraphics[trim=0pt 0pt  0pt 0pt, clip=true, width=0.50\textwidth]{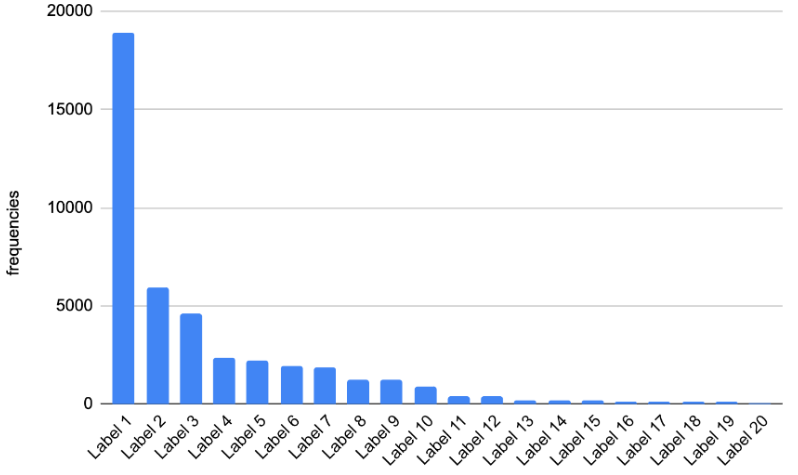}\label{fig:prop_utt_lbls}}
\caption{Proprietary dataset's conversation and utterance label distributions}
\label{fig:prop_data}

\end{figure*}

Figure \ref{fig:abcd_data} and \ref{fig:prop_data} show the conversation level and utterance level label distribution of the Repurposed ABCD and Proprietary corpora, respectively. 

\subsubsection{Model Implementation Details} \label{subsec:appx_impl_details}

\noindent\textbf{Model Implementation:}
Huggingface implementation of Reformer\footnote{\url{https://huggingface.co/docs/transformers/model_doc/reformer}}, Nystr{\"o}mformer\footnote{\url{https://huggingface.co/docs/transformers/model_doc/nystromformer}} and Longformer\footnote{\url{https://huggingface.co/docs/transformers/model_doc/longformer}}, and an open source PyTorch implementation for Performer\footnote{\url{https://github.com/lucidrains/performer-pytorch}} \citep{choromanski2021rethinking} were used in this paper. 
For the decoders, we use a single-layer classification head for both the conversationa and utterance level tasks. 
We use softmax for conversation task (multi-class classification) and sigmoid for utterance task (multi-label classification).

\noindent\textbf{Training Hyperparameters:}
The batch size was set to $16$ conversations for all training jobs across conversational datasets. 
The model was trained for $6,000$ steps on the Repurposed ABCD dataset. 
In proprietary dataset, the model was trained for $10,000$ steps.

\noindent\textbf{Model Hyperparameters:}
As we choose the model hyperparameters, we aim at maintaining a comparable number of parameters across models, with exception to the pretrained Longformer and CNN\textsubscript{Small}.
For \textbf{Longformer}, we inherit the pretrained model\footnote{\url{https://huggingface.co/allenai/longformer-base-4096}} and fine-tune it to our conversational tasks.
For \textbf{Nystr{\"o}mformer} model, we use eight attention heads and a feed forward dimension size of $3,072$.
The \textbf{Performer} model uses eight attention heads and softmax approximation hyperparameters as described in \citet{choromanski2021rethinking} instead of generalized attention parameters, as the former yielded better performance. 
For \textbf{Reformer} model, we use $2,048$ dimensional feed forward and attention mechanism alternating between local and locality-sensitive hashing in its layers.

\begin{table}[h]
    \centering
    \resizebox{2.8in}{!}{
    \begin{tabular}{c|c|c|c}
    \toprule
    & & \multicolumn{2}{c}{\#filters} \\ 
    Layer & Dilation(\textit{d}) & (CNN\textsubscript{\textit{Large}}) &  (CNN\textsubscript{\textit{Small}})\\
    \midrule
    $1$ & $1$ & $128$ & $48$ \\
    $2$ & $2$ & $256$ & $96$ \\
    $3$ & $4$ & $512$ & $192$ \\
    $4$ & $8$ & $1,024$ & $384$ \\
    \bottomrule
    \end{tabular}}
    \caption{CNN model parameters used in conversational understanding model experiments.}
    \label{table:conv_model_cnn_params}
\end{table}

For our \textbf{CNN} models, we follow strategies from \citet{bai2018empirical} for weights initialization, dilation factor and residual connections. 
In \textbf{CNN}\textsubscript{Large}, we set the kernel size $k=11$ in one stack and $k=15$ in the other, for all TCN layers.

For \textbf{CNN}\textsubscript{Small}, we scale down the CNN\textsubscript{Large} model parameters by reducing the kernel size and number of filters. 
We set the kernel size $k=5$ in one stack and $k=11$ in the other, for all layers.
The dilation factor and number of filters for each layer of both the models can be found in Table \ref{table:conv_model_cnn_params}.



\subsection{LRA benchmark experiments} \label{subsec:appx_lra_impl_details}
In this section we will present the details of different implementations and training settings that we used for evaluating the models against the LRA benchmark tasks.
\subsubsection{Training hyperparameters}
As mentioned in section \ref{subsec:lra_impl_details}, we follow the hyperparameter settings described in \citet{xiong2021nystromformer} and their implementation\footnote{\url{https://github.com/mlpen/Nystromformer}} for the LRA evaluation. Their implementation includes Full-attention, Reformer, Performer and Nystr{\"o}mformer models.

We use Adam optimizer, $\eta=1e-4$, batch size of $32$ and a linear learning rate decay strategy. 
We used a weight decay factor of $0.01$ instead of $0$ for all the models and achieved better results than reported in \citet{xiong2021nystromformer}. 


\begin{table}[h]
    \centering
    \resizebox{\columnwidth}{!}{
    \begin{tabular}{l|r|r|r}
    \toprule
    & \multicolumn{3}{c}{Training speed (\textit{steps/sec})} \\
    \midrule
    Model & Text & ListOps & Retrieval \\
    \midrule
    Full Attention & 4.8 (1.0$\times$) & 11.7 (1.0$\times$) & 2.1 (1.0$\times$) \\
    \midrule
    Reformer & 7.5 (1.6$\times$) & 9.1 (0.8$\times$) & 3.9 (1.9$\times$) \\
    Performer & 8.5 (1.8$\times$) & 10.3 (0.9$\times$) & 4.5 (2.1$\times$) \\
    Nystr{\"o}mformer & 7.9 (1.6$\times$) & 9.0 (0.8$\times$) & 3.9 (1.9$\times$) \\
    \midrule
    CNN \textit{(ours)} & \textbf{9.9} (2.1$\times$) & \textbf{11.8} (1.0$\times$) & \textbf{9.3} (4.4$\times$) \\
    \bottomrule
    \end{tabular}}
    \caption{LRA Benchmark task results - Model's Training speed}
    \label{table:appx_lra_tr_speed}
\end{table}

\subsubsection{CNN model hyperparameters} \label{subsec:appx_cnn_hyper}
Our CNN model for the \textbf{Text} task uses a similar architecture as described in Section \ref{sec:our_model}. 
We have two stacks with three TCN layers in each. 
We set \textit{k}=$9$ across layers in one stack and $k=13$ in the other. 
Dilation factor and number of filters used can be found in Table \ref{table:lra_text_listops_cnn_params}.
We use the same hyperparameters for the \textbf{ListOps} task as well.

\begin{table}[h]
    \centering
    \resizebox{1.8in}{!}{
    \begin{tabular}{c|c|c}
    \toprule
    Layer & Dilation(\textit{d}) & \#filters\\
    \midrule
    $1$ & $1$ & $8$ \\
    $2$ & $2$ & $16$ \\
    $3$ & $4$ & $32$ \\
    \bottomrule
    \end{tabular}}
    \caption{CNN model parameters used in LRA benchmark tasks - Text and ListOps.}
    \label{table:lra_text_listops_cnn_params}
\end{table}

In \textbf{Retrieval} task, we need to have a global understanding of the documents to perform well. So, instead of having two stacks of TCN layers, one for short-range and one for long-range representation as we proposed, we use only one stack (long-range) and maximize its reception field. 

To maximize the reception field, we increase the kernel size (\textit{k}). 
We experimented with $k=\{17,21,65,129,257,513\}$. 
We observed steady increase in the performance as we increased the reception field and plateaued when $k=257$ (see Table \ref{table:appx_recep_abl}). 

As the model name goes, in all layers, we set $k=17$ for CNN\textsubscript{k=17}, $k=21$ for CNN\textsubscript{k=21}, and $k=257$ for CNN\textsubscript{k=257}. Details on dilation factor and number of filters can be found in Table \ref{table:lra_retr_cnn_params}.

\begin{table}[h]
    \centering
    \resizebox{3in}{!}{
    \begin{tabular}{llcrr}
    \toprule
    Model & Acc. & FLOPs & Tr. speed & Mem. \\
    \midrule
    Full Attn. & 79.8 & 9.1 & 1.0$\times$ & 70.8 \\
    \midrule
    CNN\textsubscript{\textit{k=17}} & 72.8 & 1.0 & 4.4$\times$ & 7.4 \\
    CNN\textsubscript{\textit{k=21}} & 74.5 & 1.6 & 4.0$\times$ & 7.4 \\
    CNN\textsubscript{\textit{k=65}} & 76.7 & 2.1 & 3.2$\times$ & 7.4 \\
    CNN\textsubscript{\textit{k=129}} & 77.2 & 4.3 & 2.5$\times$ & 7.5 \\
    CNN\textsubscript{\textit{k=257}} & 79.4 & 5.9 & 1.4$\times$ & 7.6 \\
    CNN\textsubscript{\textit{k=513}} & 79.1 & 7.6 & 0.9$\times$ & 7.7 \\
    \bottomrule
    \end{tabular}}
    \caption{Ablation study. The CNN model for Retrieval task in Table \ref{table:lra_perf} has the setting \textit{k}=17 and we increase it in this study.}
    \label{table:appx_recep_abl}
\end{table}

\begin{table}[h]
    \centering
    \resizebox{2.5in}{!}{
    \begin{tabular}{c|c|c|c}
    \toprule
    Model & Layer & Dilation(\textit{d}) & \#filters\\
    \midrule
    \multirow{3}{*}{CNN\textsubscript{k=17}}
    & $1$ & $1$ & $32$ \\
    & $2$ & $2$ & $48$ \\
    \midrule
    \multirow{3}{*}{CNN\textsubscript{k=21}}
    & $1$ & $1$ & $32$ \\
    & $2$ & $2$ & $64$ \\
    \midrule
    \multirow{3}{*}{CNN\textsubscript{k=65}}
    & $1$ & $1$ & $32$ \\
    & $2$ & $2$ & $64$ \\
    \midrule
    \multirow{3}{*}{CNN\textsubscript{k=129}}
    & $1$ & $1$ & $32$ \\
    & $2$ & $2$ & $64$ \\
    \midrule
    \multirow{3}{*}{CNN\textsubscript{k=257}}
    & $1$ & $2$ & $32$ \\
    & $2$ & $4$ & $64$ \\
    \midrule
    \multirow{3}{*}{CNN\textsubscript{k=513}}
    & $1$ & $2$ & $32$ \\
    & $2$ & $4$ & $64$ \\
    \bottomrule
    \end{tabular}}
    \caption{CNN model parameters used in LRA benchmark's Retrieval task and its corresponding ablation study.}
    \label{table:lra_retr_cnn_params}
\end{table}

\begin{table}[h]
    \centering
    \resizebox{\columnwidth}{!}{
    \begin{tabular}{l|c|c|c}
    \toprule
    & \multicolumn{3}{c}{Model performance \- Acc. ($\pm${sd})} \\
    \midrule
    Models & Text & ListOps & Retrieval \\
    \midrule
    Full Attention & 65.7 ($\pm${0.2}) & \textbf{36.8} ($\pm${0.3}) & 79.8 ($\pm${1.2}) \\
    \midrule
    Reformer & 65.2 ($\pm${0.2}) & 24.9 ($\pm${4.4}) & 78.3 ($\pm${0.5}) \\
    Performer & 65.1 ($\pm${0.4}) & 28.1 ($\pm${9.9}) & 79.3 ($\pm${1.2}) \\
    Nystr{\"o}mformer & 65.8 ($\pm${0.1}) & 31.4 ($\pm${8.9}) & \textbf{80.3} ($\pm${0.7})\\
    \midrule
    CNN \textit{(ours)} & \textbf{68.0} ($\pm${0.3}) & 17.8 ($\pm${0.1}) & 72.8 ($\pm${1.2})\\
    \bottomrule
    \end{tabular}}
    \caption{LRA Benchmark task results - Model performance (Accuracy) with standard deviation}
    \label{table:appx_lra_perf}
\end{table}

\subsection{Results}
\subsubsection{Conversational Understanding - Results} \label{subsec:appx_res_std}
Table \ref{table:appx_conv_und_res_w_std} has the conversational understanding tasks model performance when trained in MTL and STL paradigms for our use case, along with the standard deviations to the F1-scores. 
We observe that the CNN's overall performance is better than the baselines in both learning paradigms, and across datasets. 
The Longformer gains the most from cross-learning between tasks.
And, not all models perform better when trained using MTL compared to STL, indicating that a model’s performance in a given learning paradigm is highly parameter or data driven. 
Additionally, we see that our CNN model has relatively lower deviation in performance across different random seeded experiments compared to the baselines.

\subsubsection{Results of LRA Benchmark} \label{subsec:appx_res_std}
In Table \ref{table:appx_lra_tr_speed} we see the models' actual training speed (steps/second) along with the relative speed from full-attention model.
Table \ref{table:appx_lra_perf} adds the standard deviation to the F1-scores reported in Table \ref{table:conv_und_results}.

\end{document}